\pdfoutput=1

\documentclass[11pt]{article}

\usepackage[final]{acl}

\usepackage{times}
\usepackage{latexsym}

\usepackage[T1]{fontenc}

\usepackage[utf8]{inputenc}

\usepackage{microtype}

\usepackage{inconsolata}

\usepackage{booktabs} 
\usepackage{graphicx}
\usepackage{lipsum}
\usepackage{tabularx}
\usepackage{enumitem} 
\usepackage{multirow}
\usepackage{pifont}

\title{Don't Buy it! Reassessing the Ad Understanding Abilities\\ of Contrastive Multimodal Models}

\author{Anna Bavaresco, Alberto Testoni, Raquel Fernández \\Institute for Logic, Language and Computation \\ University of Amsterdam \\ \texttt{ \{a.bavaresco, a.testoni, raquel.fernandez\}@uva.nl}}

\begin{document}
\maketitle
\begin{abstract}
Image-based advertisements are complex multimodal stimuli that often contain unusual visual elements and figurative language. Previous research on automatic ad understanding has reported impressive zero-shot accuracy of contrastive vision-and-language models (VLMs) on an ad-explanation retrieval task. Here, we examine the original task setup and show that contrastive VLMs can solve it by exploiting grounding heuristics. To control for this confound, we introduce TRADE, a new evaluation test set with adversarial grounded explanations. While these explanations look implausible to humans, we show that they ``fool'' four different contrastive VLMs. Our findings highlight the need for an improved operationalisation of automatic ad understanding that truly evaluates VLMs' multimodal reasoning abilities. We make our code and TRADE available at \url{https://github.com/dmg-illc/trade}.

\end{abstract}

\section{Introduction}
\begin{figure*}[t]
    \centering
    \includegraphics[width=\textwidth ]{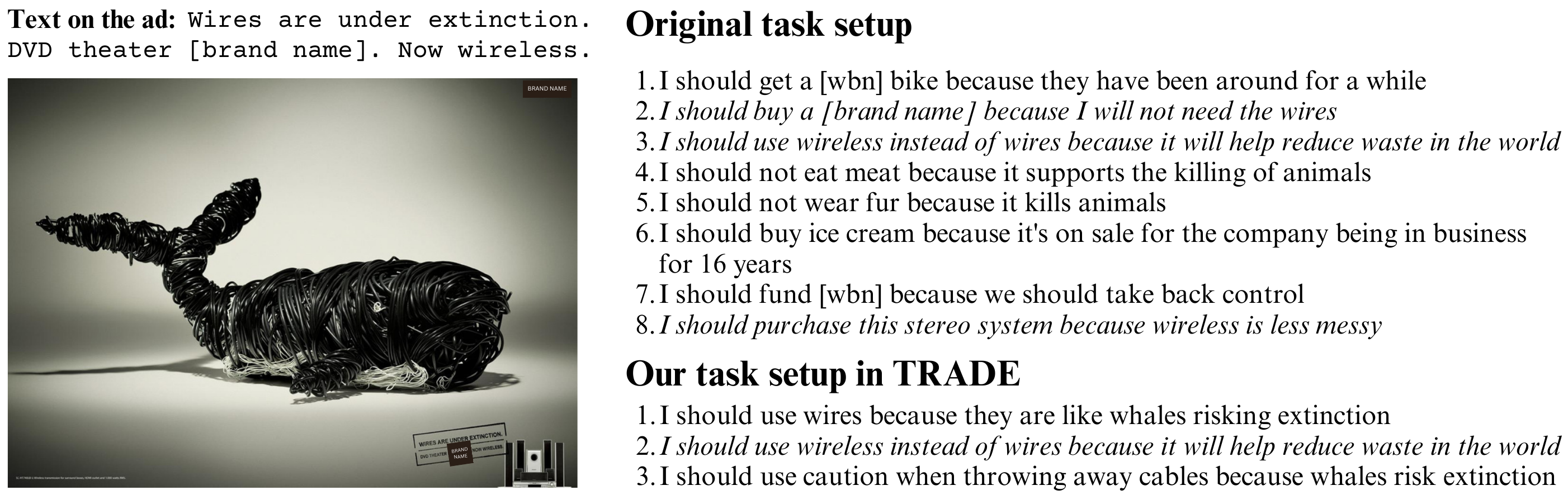}
    \caption{An example of the ad explanation retrieval task with the original setup vs.~our new setup. The matching explanations 
    are marked in italics. In the original setup, negatives are randomly sampled (5 out of 12 are shown for conciseness); in our setup, negatives are carefully curated to be textually and visually grounded in the ad but, at the same time, clearly incompatible with it. Brand names and logos are edited out in the examples present in this paper for presentation purposes, but are in fact visible in both task setups ([wbn] stands for ``wrong brand name'').}
    \label{fig:task_overview}
\end{figure*}
Image-based advertisement is not only a crucial component of marketing campaigns, but also an interesting example of sophisticated multimodal communication. Ads often feature unusual visual elements (e.g., objects that are non-photorealistic, outside of their usual context, atypical, etc.) or examples of figurative language (e.g., metaphors, allegories, play on words, etc.) designed to make a long-lasting impression on the viewer. Figure \ref{fig:task_overview} provides an example of an ad with a non-photorealistic object (a whale made of wires) that, as the text suggests, is used to convey a complex metaphorical message about the product (i.e., a wireless device). 

These elaborate uses of images and text make automatic ad understanding a challenging task requiring multiple non-trivial abilities, e.g., object detection, scene-text extraction, figurative language understanding, and complex image-text integration. Ad understanding was first proposed as a deep-learning task by \citet{hussain2017automatic}, who introduced the Pitt Ads dataset, consisting of image-based ads along with explanations capturing their underlying message (e.g., \textit{I should purchase this stereo system because wireless is less messy}). This dataset was then used in a retrieval-based challenge requiring to identify a plausible explanation for an ad 
within a set of possible candidates.\footnote{\url{https://eval.ai/web/challenges/challenge-page/86/overview}}  

Early work on this task has employed ensemble predictors \cite{hussain2017automatic, ye2018advise} and graph neural networks \cite{dey2021beyond} that were designed and trained \textit{ad hoc}. More recently, the development of large vision-and-language models (VLMs) pretrained with image-text matching (ITM) objectives has opened the possibility of performing the task in zero-shot, i.e. by using an off-the-shelf model instead of training one from scratch. Following this approach, \citet{jia-etal-2023-kafa} tested multiple VLMs (ALBEF, \citeauthor{albef}~\citeyear{albef}; CLIP, \citeauthor{clip}~\citeyear{clip}; and LiT, \citeauthor{lit}~\citeyear{lit}) 
on the task by computing image-text alignment scores between ads and their possible explanations. They observed an excellent zero-shot performance for all models, documenting an accuracy of 95.2\% for CLIP. 

While the results reported by \citet{jia-etal-2023-kafa} seem to suggest that the tested models 
developed the reasoning abilities necessary to succeed at ad understanding, we note that this conclusion is in contrast with a great deal of existing work. Extensive research investigating whether VLMs develop reasoning skills as a result of their contrastive ITM pretraining has exposed several weaknesses of these models. They have been shown to be limited in their abilities to identify noun mismatches in image captions \cite{shekhar-etal-2017-foil}, reason compositionally \cite{winoground}, capture spatial relations \cite{vsr}, understand verbs (instead of just nouns) \cite{svo_probing},  and handle various linguistic phenomena \cite{valse} and basic constructions \cite{chen-etal-2023-bla}.


Importantly, this line of work focused on a set of traditional visuo-linguistic tasks but not specifically on ad understanding. Here, we ask whether the performance previously documented on the Pitt Ads dataset reflects genuine understanding abilities or is driven by simpler heuristics.
We conduct a thorough analysis of the evaluation setup originally proposed to test ad understanding and reveal that it has key flaws, which allow models to exploit grounding heuristics. 
We introduce a new test set, TRADE (\textit{TRuly ADversarial ad understanding Evaluation}), which controls for the identified issues. 
Our experiments show that several contrastive models tested zero-shot, including CLIP, perform at chance level on TRADE, while humans excel at the task. 
More generally, our findings highlight the need to better operationalise ad understanding in order to obtain reliable assessments of VLMs' multimodal reasoning abilities.

\section{A Closer Look at the Evaluation Setup}
\label{sec:setup_issues}
The Pitt Ads dataset\footnote{\url{https://people.cs.pitt.edu/~kovashka/ads/}} by 
\citet{hussain2017automatic} 
consists of $64832$ ads, each annotated with 3 explanations in English written by 3 different expert annotators. These explanations (in the form \textit{I should $\langle$action$\rangle$ because $\langle$reason$\rangle$}) aim at capturing the persuasive message behind the ads. While explanations may be subjective, the intuition behind the image-to-text retrieval task proposed along with the dataset is that a model which can understand ads should be able to match them with a plausible explanation. Specifically, each ad is paired with 15 messages, 3 positives corresponding to the annotations for that ad and 12 negatives randomly sampled from annotations for different ads. Figure \ref{fig:task_overview} provides an overview of the task setup. 

Previous work has hinted at possible limitations of the evaluation setup. \citet{bert-ads} observed a significant overlap between the text present in the ad and the matching explanations and noticed this was ``a major discriminating factor'' that their fine-tuned BERT model could exploit. Similarly, \citet{jia-etal-2023-kafa} pointed out that the candidate set lacks ``hard negatives'' and proposed to increase the set size, but could not provide a solution ensuring the negatives were actually hard. 

We conduct a quantitative analysis on the original evaluation setup to uncover potential shortcuts that VLMs may be exploiting to solve the task. We hypothesise that the models may take advantage of two factors that do not necessarily reflect ad understanding: simple relationships between (1) the candidate explanations and the text present in the ad (i.e., the degree of \textit{textual grounding} of the explanations) and (2) the entities mentioned in the explanations and those depicted in the image (their degree of \textit{visual grounding}).
To test our hypotheses, we define several visual- and textual-grounding scores 
and check whether they correlate with the CLIP-based alignment score used by \citet{jia-etal-2023-kafa} to retrieve the ad explanations.\footnote{More details on the scores can be found in Appendix~\ref{sec:appendix_grounding_scores}.}

\paragraph{Textual-grounding scores} 
are computed between candidate explanations and the text extracted from the ad with Optical Character Recognition (OCR). We calculate (1) \textit{text overlap} as the proportion of content-word lemmas   
from the explanation that are also present in the OCR-extracted text,  and (2) \textit{text similarity} as the cosine similarity between a sentence-level embedding of the explanation and that of the OCR-extracted text, derived with MPNet \cite{mpnet}.

\paragraph{Visual-grounding scores} 
 include (1) \textit{object mention} as the proportion of nouns in the candidate explanation that are present in a set of objects we automatically extracted from the image by a ResNet50 model \cite{resnet}, and (2) \textit{caption similarity} as the cosine similarity between the sentence-level embedding of the candidate explanation and the embedding of the ad caption we obtained with BLIP-2 \cite{blip2}. Our motivation for examining both detected objects and generated captions is driven by the observation that they capture complementary information. More specifically, while detected objects are not mediated by language models, they may often incorporate non-salient objects that people would unlikely mention when describing a picture or contain lexical choices that differ from the human ones. On the other hand, generated captions refer to objects in a more human-like way but, at the same time, may contain hallucinations due to linguistic priors.

We compute the grounding scores and CLIP's alignment score for the test split of the Pitt Ads dataset, consisting of $12805$ samples. As hypothesised, we observe a positive correlation between all our grounding scores and CLIP's alignment score. All the Spearman's correlation coefficients are significant ($p\ll0.001$) and range from $0.14$ and $0.61$ (see Appendix~\ref{sec:appendix_grounding_scores} for details). In addition, as shown in Table~\ref{tab:grounding-scores} (left), we find that in the original setup the matching explanations are significantly more grounded than the non-matching explanations for each ad. While the elements (OCR text, objects, captions) detected by other models are not necessarily the same as those identified by CLIP,  these results suggest that reasonably similar information is indirectly extracted by CLIP and exploited to solve the ad-understanding task. This finding also agrees with results from previous work showing that CLIP develops OCR capabilities and can successfully 
classify objects \cite{clip}.

Overall, these results indicate that the original evaluation setup is flawed and that the outstanding zero-shot performance obtained by VLMs on the retrieval task may be due to simple image-text alignment.

\begin{table}[]
\resizebox{\columnwidth}{!}{
    \centering
    \begin{tabular}{rll@{\ }l@{}}\toprule
                            & \multicolumn{3}{l}{\bf original setup}\\\midrule
                            & \bf Pos  & \bf Neg &  \\
    \textit{text overlap}   & 0.21 & 0.03& * \\
    \textit{text similarity}& 0.40 & 0.12& * \\
    \textit{object mention} & 0.03 & 0.01& * \\
    \textit{caption similarity}   & 0.32 & 0.11& * \\\bottomrule
    \end{tabular} \ \ \ 
    \begin{tabular}{ll@{\ }l}\toprule
       \multicolumn{3}{c}{\bf TRADE}\\\midrule
        \bf Pos & \bf Neg & \\
          0.27 & 0.31 & * \\
          0.44 & 0.42 & \\
          0.02 & 0.04 & \\
          0.34 & 0.35 \\\bottomrule
    \end{tabular}
}
    \caption{Average textual- and visual-grounding scores of the matching (Pos) and non-matching (Neg) explanations in the original evaluation setup and in TRADE; statistically significant differences between Pos and Neg marked with * ($p \ll 0.001$, two-sample t-test).}
    \label{tab:grounding-scores}
\end{table}

\section{TRADE: A New Adversarial Test Set}
\label{sec:trade}

To test the extent to which VLMs capture elaborate visuo-linguistic relationships present in image-based ads beyond image-text alignment, we develop TRADE (\textit{TRuly ADversarial ad understanding Evaluation}), a new diagnostic test set with adversarial negative explanations.
TRADE  consists of 300 randomly selected ads from the Pitt Ads dataset, each associated with 3 options (1 positive and 2 negatives).
Concretely, for each of these ads, we randomly select one valid explanation from the available annotations and create two adversarial negative explanations---see Figures~\ref{fig:task_overview} and \ref{fig:performance_example} for examples (more examples in Appendix~\ref{sec:appendix_examples}). The adversarial explanations were created by 4 expert annotators who were instructed to do their best to come up with non-plausible explanations that nevertheless mention objects and fragments of text present in the image. Annotators were also asked to approximately match the length of the positive explanation when writing these adversarial sentences. 
Appendix~\ref{sec:appendix_collection_negatives} contains more details about the creation of the adversarial negatives, including the guidelines provided to the annotators.

We validate TRADE in two ways. First, we compute the textual- and visual-grounding scores introduced in Section~\ref{sec:setup_issues}. This shows that in TRADE the gap between positive and negative explanations is radically reduced compared to the original setup, as can be seen in Table~\ref{tab:grounding-scores} (right). 
Second, we confirm that humans are not affected by the high level of grounding of both positive and negative examples and are able to identify the plausible explanation in the TRADE samples with an accuracy of 94\%.\footnote{Each of the 300 samples was annotated by two annotators external to the project; more details available in Appendix \ref{sec:appendix_human_acc}.}

To allow for a direct comparison with an evaluation setup with random negatives, akin to the original task setup, we also create TRADE-control: a version of TRADE where the two negative explanations per ad are randomly sampled from the explanations for other ads. TRADE-control includes 10 versions created with different random samplings.

TRADE and TRADE-control are publicly available at \url{https://github.com/dmg-illc/trade} 
under a Creative Commons Attribution 4.0 International (CC-BY) license.


\section{Experiments}
\label{sec:exp}

We use TRADE to test four contrastive pretrained VLMs zero-shot. 
Three of these models (CLIP, \citeauthor{clip}~\citeyear{clip}; ALBEF, \citeauthor{albef}~\citeyear{albef}; and LiT, \citeauthor{lit}~\citeyear{lit}) have been shown to achieve high zero-shot performance on the original task setup \cite{jia-etal-2023-kafa}. Here we challenge them with TRADE and consider an additional model (ALIGN, \citeauthor{align}~\citeyear{align}).


\subsection{Models and Setup}
 Except for ALBEF, all the models we test encode visual and textual inputs separately and are pretrained with an image-text matching objective. ALBEF has an additional multimodal module, but here we only use its unimodal encoders, which are also pretrained contrastively. A more detailed overview of these VLMs is reported in Appendix \ref{sec:appendix_model_overview}.

All four models allow for the computation of an image-text alignment score, here defined as the dot product between the normalized image embedding and the text embedding of each candidate explanation. As in previous work \cite{jia-etal-2023-kafa}, we evaluate the models by computing 
alignment scores for every ad-explanation pair and consider the explanation yielding the highest alignment score as the model's retrieved option. We report average accuracy, as (mean) rank is not very informative with only 3 candidates.

\subsection{Results}

\begin{figure*}
    \centering
    \begin{tabularx}{\textwidth}{@{}p{.28\textwidth}Xp{.15\textwidth}} \toprule
      \textbf{Ad} & \textbf{TRADE explanations} & \textbf{Chosen by}\\ \midrule
      \multirow{ 3}{*}{\includegraphics[width=0.27\textwidth]{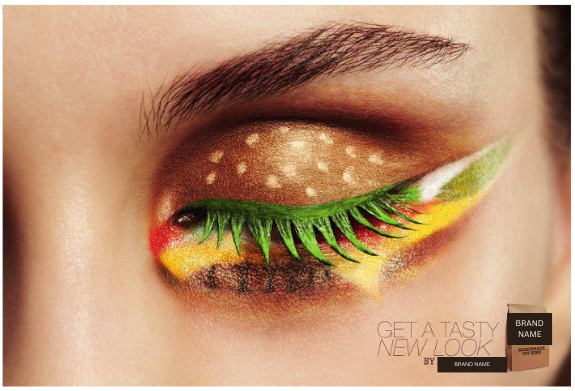}} & {1. \emph{I should go to [brand name] not only does their \hspace*{0.3cm} food taste great but it also looks good.}}  & Human \\ 
       & {2. I should go to [brand name] because my eyelashes \hspace*{0.3cm} need a new look.} & CLIP, ALIGN  \\
      & {3. I should go to [brand name] because tasty burgers \hspace*{0.3cm} must look like these eyelashes.} & ALBEF, LiT \\ \bottomrule
  
    \end{tabularx}
    \caption{Ad explanations selected by human annotators vs.\ our tested models for one instance from TRADE. Italic indicates the \emph{matching explanation}. Brands and logos are edited out in the paper examples for presentation purposes but are visible to models and human annotators.}
    \label{fig:performance_example}
\end{figure*}

\begin{table}[t]
\resizebox{\columnwidth}{!}{
\centering
\begin{tabular}{lcc}\toprule

\textbf{Model} & \textbf{TRADE} & \textbf{control}\\
\midrule
CLIP (ViT-L/14@336px) & 0.34 & 0.98 (0.01)  \\
ALIGN (base) & 0.28 & 0.97 (0.01) \\
LiT (L16L)  & 0.31 &  0.82 (0.02)\\
ALBEF (ft. on Flickr30k) & 0.33 & 0.88 (0.01) \\\bottomrule
\end{tabular}
}
\caption{Average accuracy on TRADE vs.~TRADE-control. The TRADE-control values are averages over 10 random samples, with standard deviation in brackets.}
\label{tab:model_performance}
\end{table}

Table~\ref{tab:model_performance} shows the performance of the models on 
TRADE and TRADE-control. 
All models achieve an accuracy higher than 80\% in the control condition, with CLIP reaching 98\%. However, 
the performance of all models in the adversarial setting---where humans achieve 94\% accuracy, cf.~Section~\ref{sec:trade}---nears chance level, i.e., 33\%. Figure \ref{fig:performance_example} provides an example of model- and human-chosen ad explanations on a TRADE instance.
These results provide compelling evidence that the evaluated VLMs rely on visual and textual grounding when retrieving ad explanations. As a result, they can achieve excellent accuracy in an evaluation setting where negatives are poorly grounded, but are easily ``fooled'' by grounded adversarial distractors that are extremely easy for humans to discard.

\begin{figure}
    \centering
    \includegraphics[width=0.9\columnwidth]{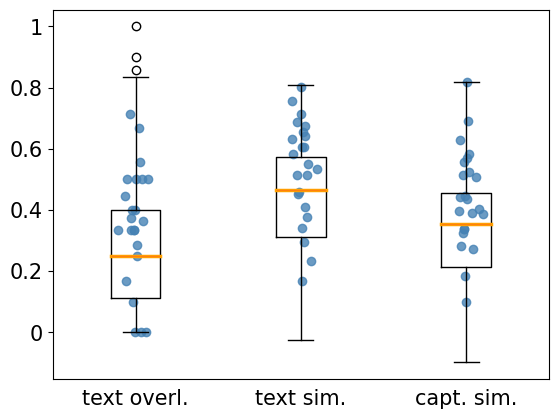}
    \caption{Boxplots summarizing the distribution of grounding scores computed for positive explanations in TRADE. The blue dots indicate the scores for the positive explanations correctly selected by all VLMs. The \textit{object mention} score is not included because its median coincides with the quartiles.}
    \label{fig:boxplots}
\end{figure}

To get more insight into the models' performance, we examine their predictions and observe that,
while all models perform equally poorly on TRADE, there are 23 samples (8\% of the dataset) for which the four models succeed at identifying the target explanation. 
An analysis of the explanations correctly retrieved by all models reveals that most of them exhibit grounding scores that are higher than the average scores for matching explanations. Figure~\ref{fig:boxplots} visualises this finding.


\section{Conclusions}

Our work exposes key limitations of the evaluation setup that was previously used to benchmark VLM's ad understanding abilities. We introduce a new adversarial test set (TRADE) that controls for the identified issues and show that, while humans excel, contrastive VLMs perform at chance level on TRADE. 
This result has the following implications.

First, it shows that, when processing image-based ads,
contrastive VLMs are strongly biased towards textually and visually grounded explanations, regardless of their plausibility. This is in agreement with previous work 
\cite{svo_probing, valse, winoground, vsr, chen-etal-2023-bla} and points to the need to use caution when interpreting models' zero-shot accuracy on ``naturalistic'' (i.e., non-adversarial) setups as proof that they develop sophisticated reasoning abilities via pretraining.

Second, our work highlights issues with the current retrieval-based operationalization of ad understanding as a task to evaluate VLM's multimodal reasoning abilities. We emphasise that TRADE's aim is  
to control for a confound---the grounding gap between positives and negatives---that we identified as crucial when testing a specific type of VLMs, i.e., those pretrained with an ITM objective. However, defining which abilities are necessary to conclude that a model developed a good ``understanding'' of image-based ads and designing a task that truly evaluates them remain open issues for future research. 

\section*{Limitations}

The current study and previous work have operationalised ad understanding as an ad-explanation retrieval task. In particular, we have focused on testing contrastive pretrained VLMs zero-shot on this task. Consequently, the question of whether VLMs trained or finetuned on the Pitt Ads dataset would be more robust against our adversarial explanations remains open and could be investigated in the future. Nevertheless, we emphasize that the retrieval-based setup has limitations (e.g., the impossibility of providing task-specific instructions to the models) and may not be the most appropriate to evaluate VLM's ad understanding skills and their multimodal reasoning abilities more generally. An interesting direction for future research could be to formulate the task differently, e.g., as a generative task.
This would solve some issues of the retrieval-based setup, but also posit novel challenges, such as identifying the most effective prompt and defining meaningful protocols to evaluate the generated explanations. 

On a methodological note, we highlight that visual and textual alignment are complex constructs that encompass different aspects and can be analysed at different levels of granularity. Therefore, we do not intend our grounding scores as precise and comprehensive metrics, but simply as indicators that can reflect general trends.

\section*{Ethical Considerations}

TRADE does not introduce new ad-images, but simply links to the existing Pitt Ads dataset along with the set of adversarial explanations we have created. However, it is worth emphasizing that the ads present in Pitt Ads were originally collected by querying Google Images.
This posits two ethical concerns. 

First, offensive/harmful content or stereotypes may be present in the images, as already pointed out by \citet{jia-etal-2023-kafa}. To minimise this potential problem when developing TRADE, we made sure the annotators who created our adversarial explanations had the possibility of flagging ads that they deemed inappropriate (they did so a couple of times). However, we cannot fully guarantee that the ad images used in TRADE are completely free from harmful content. As for the adversarial distractors created for TRADE, we have not systematically examined all of them manually to make sure they do not contain harmful content, but we believe this is very unlikely given the guidelines and the fact that they were created in a very controlled setting partially by us and partially by close colleagues. 

The second concern is about the license of the images. The Pitt Ads dataset was released without a license and the curators do not clarify whether the images are copyrighted or not. 

Finally, we note that our study does not take into account the personal and cultural factors which may play a substantial role in people's perception of ads or in the values they associate with certain products. Although TRADE includes only one matching explanation for each ad, we emphasize that we do not intend this as a ``ground truth''. We hope that future research on automatic ad understanding will adopt evaluation protocols that reflect a diverse set of possible interpretations.   


\section*{Acknowledgements}
We warmly thank the current members and alumni of the Dialogue Modelling Group (DMG) from the University of Amsterdam for the support they provided at different stages of this project. Special thanks are due to Sandro Pezzelle, who suggested the name `TRADE' for our introduced dataset. Our heartfelt gratitude also goes to the colleagues who assisted us with the creation of TRADE negatives and to the colleagues, friends and partners who kindly volunteered to participate in our experiment to assess human performance on TRADE. The present work was funded by the European Research Council under the European Union’s Horizon 2020 research and innovation programme (grant agreement No.\ 819455). 

\bibliography{acl_latex}

\begin{thebibliography}{21}
\expandafter\ifx\csname natexlab\endcsname\relax\def\natexlab#1{#1}\fi

\bibitem[{Chen et~al.(2023)Chen, Fern{\'a}ndez, and Pezzelle}]{chen-etal-2023-bla}
Xinyi Chen, Raquel Fern{\'a}ndez, and Sandro Pezzelle. 2023.
\newblock \href {https://doi.org/10.18653/v1/2023.emnlp-main.356} {The {BLA} benchmark: Investigating basic language abilities of pre-trained multimodal models}.
\newblock In \emph{Proceedings of the 2023 Conference on Empirical Methods in Natural Language Processing}, pages 5817--5830, Singapore. Association for Computational Linguistics.

\bibitem[{Dey et~al.(2021)Dey, Ghosh, Valveny, and Harit}]{dey2021beyond}
Arka~Ujjal Dey, Suman~K Ghosh, Ernest Valveny, and Gaurav Harit. 2021.
\newblock Beyond visual semantics: Exploring the role of scene text in image understanding.
\newblock \emph{Pattern Recognition Letters}, 149:164--171.

\bibitem[{Goh et~al.(2021)Goh, Cammarata, Voss, Carter, Petrov, Schubert, Radford, and Olah}]{clipneurons}
Gabriel Goh, Nick Cammarata, Chelsea Voss, Shan Carter, Michael Petrov, Ludwig Schubert, Alec Radford, and Chris Olah. 2021.
\newblock Multimodal neurons in artificial neural networks.
\newblock \emph{Distill}, 6(3):e30.

\bibitem[{He et~al.(2016)He, Zhang, Ren, and Sun}]{resnet}
Kaiming He, Xiangyu Zhang, Shaoqing Ren, and Jian Sun. 2016.
\newblock \href {https://doi.org/10.1109/CVPR.2016.90} {Deep residual learning for image recognition}.
\newblock In \emph{2016 IEEE Conference on Computer Vision and Pattern Recognition (CVPR)}, pages 770--778.

\bibitem[{Hendricks and Nematzadeh(2021)}]{svo_probing}
Lisa~Anne Hendricks and Aida Nematzadeh. 2021.
\newblock \href {https://doi.org/10.18653/v1/2021.findings-acl.318} {Probing image-language transformers for verb understanding}.
\newblock In \emph{Findings of the Association for Computational Linguistics: ACL-IJCNLP 2021}, pages 3635--3644, Online. Association for Computational Linguistics.

\bibitem[{Hussain et~al.(2017)Hussain, Zhang, Zhang, Ye, Thomas, Agha, Ong, and Kovashka}]{hussain2017automatic}
Zaeem Hussain, Mingda Zhang, Xiaozhong Zhang, Keren Ye, Christopher Thomas, Zuha Agha, Nathan Ong, and Adriana Kovashka. 2017.
\newblock Automatic understanding of image and video advertisements.
\newblock In \emph{Proceedings of the IEEE Conference on Computer Vision and Pattern Recognition (CVPR)}, pages 1705--1715.

\bibitem[{Jia et~al.(2021)Jia, Yang, Xia, Chen, Parekh, Pham, Le, Sung, Li, and Duerig}]{align}
Chao Jia, Yinfei Yang, Ye~Xia, Yi-Ting Chen, Zarana Parekh, Hieu Pham, Quoc Le, Yun-Hsuan Sung, Zhen Li, and Tom Duerig. 2021.
\newblock \href {https://proceedings.mlr.press/v139/jia21b.html} {Scaling up visual and vision-language representation learning with noisy text supervision}.
\newblock In \emph{Proceedings of the 38th International Conference on Machine Learning}, volume 139, pages 4904--4916.

\bibitem[{Jia et~al.(2023)Jia, Narayana, Akula, Pruthi, Su, Basu, and Jampani}]{jia-etal-2023-kafa}
Zhiwei Jia, Pradyumna Narayana, Arjun Akula, Garima Pruthi, Hao Su, Sugato Basu, and Varun Jampani. 2023.
\newblock \href {https://doi.org/10.18653/v1/2023.acl-industry.74} {{KAFA}: Rethinking image ad understanding with knowledge-augmented feature adaptation of vision-language models}.
\newblock In \emph{Proceedings of the 61st Annual Meeting of the Association for Computational Linguistics (Volume 5: Industry Track)}, pages 772--785, Toronto, Canada. Association for Computational Linguistics.

\bibitem[{Kalra et~al.(2020)Kalra, Kurma, Vadakkeeveetil~Sreelatha, Patwardhan, and Karande}]{bert-ads}
Kanika Kalra, Bhargav Kurma, Silpa Vadakkeeveetil~Sreelatha, Manasi Patwardhan, and Shirish Karande. 2020.
\newblock \href {https://doi.org/10.18653/v1/2020.acl-main.674} {Understanding advertisements with {BERT}}.
\newblock In \emph{Proceedings of the 58th Annual Meeting of the Association for Computational Linguistics}, pages 7542--7547, Online. Association for Computational Linguistics.

\bibitem[{Li et~al.(2023)Li, Li, Savarese, and Hoi}]{blip2}
Junnan Li, Dongxu Li, Silvio Savarese, and Steven Hoi. 2023.
\newblock {BLIP-2}: bootstrapping language-image pre-training with frozen image encoders and large language models.
\newblock In \emph{Proceedings of the 40th International Conference on Machine Learning (ICML)}.

\bibitem[{Li et~al.(2021)Li, Selvaraju, Gotmare, Joty, Xiong, and Hoi}]{albef}
Junnan Li, Ramprasaath Selvaraju, Akhilesh Gotmare, Shafiq Joty, Caiming Xiong, and Steven Chu~Hong Hoi. 2021.
\newblock \href {https://proceedings.neurips.cc/paper_files/paper/2021/file/505259756244493872b7709a8a01b536-Paper.pdf} {Align before fuse: Vision and language representation learning with momentum distillation}.
\newblock In \emph{Advances in Neural Information Processing Systems}, volume~34, pages 9694--9705. Curran Associates, Inc.

\bibitem[{Lin et~al.(2014)Lin, Maire, Belongie, Hays, Perona, Ramanan, Doll{\'a}r, and Zitnick}]{cocodataset}
Tsung-Yi Lin, Michael Maire, Serge Belongie, James Hays, Pietro Perona, Deva Ramanan, Piotr Doll{\'a}r, and C~Lawrence Zitnick. 2014.
\newblock Microsoft coco: Common objects in context.
\newblock In \emph{Proceedings of the 13th European Conference on Computer Vision (ECCV)}, pages 740--755. Springer.

\bibitem[{Liu et~al.(2023)Liu, Emerson, and Collier}]{vsr}
Fangyu Liu, Guy Emerson, and Nigel Collier. 2023.
\newblock \href {https://doi.org/10.1162/tacl_a_00566} {Visual spatial reasoning}.
\newblock \emph{Transactions of the Association for Computational Linguistics}, 11:635--651.

\bibitem[{Parcalabescu et~al.(2022)Parcalabescu, Cafagna, Muradjan, Frank, Calixto, and Gatt}]{valse}
Letitia Parcalabescu, Michele Cafagna, Lilitta Muradjan, Anette Frank, Iacer Calixto, and Albert Gatt. 2022.
\newblock \href {https://doi.org/10.18653/v1/2022.acl-long.567} {{VALSE}: A task-independent benchmark for vision and language models centered on linguistic phenomena}.
\newblock In \emph{Proceedings of the 60th Annual Meeting of the Association for Computational Linguistics (Volume 1: Long Papers)}, pages 8253--8280, Dublin, Ireland. Association for Computational Linguistics.

\bibitem[{Radford et~al.(2021)Radford, Kim, Hallacy, Ramesh, Goh, Agarwal, Sastry, Askell, Mishkin, Clark, Krueger, and Sutskever}]{clip}
Alec Radford, Jong~Wook Kim, Chris Hallacy, Aditya Ramesh, Gabriel Goh, Sandhini Agarwal, Girish Sastry, Amanda Askell, Pamela Mishkin, Jack Clark, Gretchen Krueger, and Ilya Sutskever. 2021.
\newblock \href {https://proceedings.mlr.press/v139/radford21a.html} {Learning transferable visual models from natural language supervision}.
\newblock In \emph{Proceedings of the 38th International Conference on Machine Learning}, volume 139, pages 8748--8763.

\bibitem[{Savchenko et~al.(2020)Savchenko, Alekseev, Kwon, Tutubalina, Myasnikov, and Nikolenko}]{adlingua}
Andrey Savchenko, Anton Alekseev, Sejeong Kwon, Elena Tutubalina, Evgeny Myasnikov, and Sergey Nikolenko. 2020.
\newblock \href {https://doi.org/10.18653/v1/2020.coling-main.171} {Ad lingua: Text classification improves symbolism prediction in image advertisements}.
\newblock In \emph{Proceedings of the 28th International Conference on Computational Linguistics}, pages 1886--1892, Barcelona, Spain (Online). International Committee on Computational Linguistics.

\bibitem[{Shekhar et~al.(2017)Shekhar, Pezzelle, Klimovich, Herbelot, Nabi, Sangineto, and Bernardi}]{shekhar-etal-2017-foil}
Ravi Shekhar, Sandro Pezzelle, Yauhen Klimovich, Aur{\'e}lie Herbelot, Moin Nabi, Enver Sangineto, and Raffaella Bernardi. 2017.
\newblock \href {https://doi.org/10.18653/v1/P17-1024} {{FOIL} it! find one mismatch between image and language caption}.
\newblock In \emph{Proceedings of the 55th Annual Meeting of the Association for Computational Linguistics (Volume 1: Long Papers)}, pages 255--265, Vancouver, Canada. Association for Computational Linguistics.

\bibitem[{Song et~al.(2020)Song, Tan, Qin, Lu, and Liu}]{mpnet}
Kaitao Song, Xu~Tan, Tao Qin, Jianfeng Lu, and Tie-Yan Liu. 2020.
\newblock Mpnet: masked and permuted pre-training for language understanding.
\newblock In \emph{Proceedings of the 34th International Conference on Neural Information Processing Systems (NeurIPS)}, Red Hook, NY, USA. Curran Associates Inc.

\bibitem[{Thrush et~al.(2022)Thrush, Jiang, Bartolo, Singh, Williams, Kiela, and Ross}]{winoground}
Tristan Thrush, Ryan Jiang, Max Bartolo, Amanpreet Singh, Adina Williams, Douwe Kiela, and Candace Ross. 2022.
\newblock Winoground: Probing vision and language models for visio-linguistic compositionality.
\newblock In \emph{Proceedings of the IEEE/CVF Conference on Computer Vision and Pattern Recognition (CVPR)}, pages 5238--5248.

\bibitem[{Ye and Kovashka(2018)}]{ye2018advise}
Keren Ye and Adriana Kovashka. 2018.
\newblock Advise: Symbolism and external knowledge for decoding advertisements.
\newblock In \emph{Proceedings of the European Conference on Computer Vision (ECCV)}, pages 837--855.

\bibitem[{Zhai et~al.(2022)Zhai, Wang, Mustafa, Steiner, Keysers, Kolesnikov, and Beyer}]{lit}
Xiaohua Zhai, Xiao Wang, Basil Mustafa, Andreas Steiner, Daniel Keysers, Alexander Kolesnikov, and Lucas Beyer. 2022.
\newblock Lit: Zero-shot transfer with locked-image text tuning.
\newblock In \emph{Proceedings of the IEEE/CVF Conference on Computer Vision and Pattern Recognition (CVPR)}, pages 18123--18133.

\end{thebibliography}


\appendix

\section*{Appendix}

\section{Grounding Scores}
\label{sec:appendix_grounding_scores}

\paragraph{Textual Grounding}

The textual grounding scores were computed between candidate explanations and the ad OCR-extracted text output by Google Vision API\footnote{\url{https://cloud.google.com/vision/docs/ocr}} and made publicly available\footnote{\url{https://figshare.com/articles/dataset/OCR_results/6682709}} by the authors of \citet{adlingua}. OCR text was present for $12304$ ad-images of the test set and $294$ images from TRADE. 
The \textit{text overlap} score was computed as the proportion of words from the candidate explanation that were also present in the OCR-extracted text. Before computing the overlap, we lemmatized the text and removed stop-words. These preprocessing steps were performed with the NLTK\footnote{\url{https://www.nltk.org/}} package. 

The \textit{text similarity} score was defined as the cosine similarity between the embedding of the explanation and that of the OCR-extracted text. The embeddings were obtained using the Sentence Transformers\footnote{\url{https://github.com/UKPLab/sentence-transformers?tab=readme-ov-file}} framework. Specifically, we used an MPNet \cite{mpnet} pretrained model, which was indicated as the best-performing one. 

\paragraph{Visual Grounding}
To compute the visual grounding scores, we considered two sources of visual information: the objects identified by an object detector, and the ad-image captions. Our object detector was a ResNet50 model \cite{resnet} pretrained on MS COCO \cite{cocodataset}. We used the implementation from the Detectron2\footnote{\url{https://github.com/facebookresearch/detectron2}} framework by Facebook. The model detected an average of $3.74\pm3.85$ objects from the Pitt Ads dataset test split and $3.51\pm3.38$ from TRADE. At least one object was detected on $11351$ images from the Pitt Ads dataset test split and on all the ads from TRADE ($300$). Ad captions were obtained using BLIP2 \cite{blip2} with OPT 2.7B as language decoder. BLIP2 was used in its Hugging Face implementation.\footnote{\url{https://huggingface.co/docs/transformers/model_doc/blip-2}}

The \textit{object mention} score was computed as the lemmatized nouns in the AR statement that were part of the set of detected objects. The \textit{caption similarity} score, on the other hand, was defined similarly to the \textit{text similarity} score, with the caption in place of the OCR-extracted text. 

\paragraph{Correlation with CLIP's alignment scores}

We computed Spearman correlations between all the grounding scores and the CLIP-alignment scores for both the Pitt Ads test set and TRADE. 

All results are summarized in Tables \ref{tab:vis_tex_grounding} and \ref{tab:vis_tex_grounding_ours}.


\begin{table*}[]
    \centering
    \begin{tabular}{llllll} \toprule
       & \textbf{Pos} & \textbf{Neg} & \textbf{CLIP-pos} & \textbf{CLIP-neg} & \textbf{Corr}\\ \midrule
     \textit{text overlap} & 0.21 & 0.03 & 23.78 & 12.66 &  0.28 \\
      \textit{text similarity} & 0.4 & 0.12 & 23.78 & 12.66 & 0.61\\
      \textit{object mention} & 0.03 & 0.01 & 23.72 & 12.74 & 0.14\\
      \textit{caption similarity} & 0.32 & 0.11 & 23.72  & 12.68 & 0.53\\\bottomrule
    \end{tabular}
    \caption{Grounding scores and CLIP-alignment scores for matching (positives) and non-matching (negatives) explanations from the original test set. Two-sample t-tests indicate that all differences between positives and negatives are statistically significant ($p\ll0.001$). The right-most column reports the Spearman correlations between aggregated (including both positives and negatives) grounding scores and the corresponding CLIP-alignment scores. All the correlation values are statistically significant ($p\ll0.001$).}
    \label{tab:vis_tex_grounding}
\end{table*}

\begin{table*}[]
    \centering
    \begin{tabular}{llllll} \toprule
       & \textbf{Pos} & \textbf{Neg}  & \textbf{CLIP-pos}  & \textbf{CLIP-neg}  & \textbf{Corr}  \\ \midrule
      \textit{text overlap} & 0.27 & 0.31  & 24.87 & 24.42 & 0.22 ($p=0$) \\
      \textit{text similarity} & 0.44 & 0.42 & 24.87 & 24.42 & 0.41 ($p=0$)\\
      \textit{object mention} & 0.02 & 0.04  & 24.84 & 24.39 &  0.04 ($p=0.22$)\\
      \textit{caption similarity} & 0.34 & 0.35 & 24.84 & 24.39 & 0.3 ($p=0$) \\\bottomrule
    \end{tabular}
    \caption{Grounding scores and CLIP-alignment scores for matching (positives) and non-matching (negatives) explanations from TRADE. With the exception of \textit{text overlap}, the differences between grounding scores are not statistically significant ($p\ll0.001$). All the differences between positive and negative CLIP-alignment scores are also non-significant.}
    \label{tab:vis_tex_grounding_ours}
\end{table*}

\section{Creating TRADE}
\label{sec:appendix_collection_negatives}

The adversarial negatives were designed by two of the authors and two internal collaborators who volunteered for the task and are all proficient in English. Due to the complexity of this annotation task, we deemed it not suitable for crowdsourcing. The instructions given to the annotators were the following:

\begin{enumerate}[leftmargin=12pt,itemsep=0pt] \small
    \item The sentence should be \underline{inconsistent} with the image, meaning that it should not be a valid answer to the question “What should you do, according to this ad?”. Keep in mind that the answer should be \underline{patently wrong}, i.e. it should require very little thinking to figure out it does not match the message of the ad.
    \item The sentence should be in the form “I should [action] because [reason]”
    \item The verb you use after “should” should be the same as the one from the right sentence. For example, if the right sentence starts with “I should buy”, your wrong annotation cannot start with “I should fly”
    \item The sentence should be as grounded as possible, meaning that you should avoid mentioning objects/words that are not present in the ad as much as you can. Please keep this in mind, \underline{it is very important}!
    \item If possible, privilege \underline{salient} visual elements over non-salient ones. More concretely, try to mention large writings instead of small ones, and big foreground objects instead of small background ones.
    \item When describing visual objects, try to \underline{be efficient} instead of verbose. For example, if an ad depicts a famous man (say, Mr. X) driving a car of a specific brand (say, Brand Y), you should write something like “I should buy Mr. X because he drives a cool Brand Y car” instead of “I should buy a man with short hair and sunglasses because he drives a red four-wheeled vehicle”
    \item Please \underline{avoid extra-long sentences}. Your wrong answers should be approximately the same length as the correct ones. You don’t need to be as strict as to count the exact number of words but try to avoid large mismatches (e.g. correct answer being not even one-line long and wrong answer being two lines)
    \item Only include the name of \underline{brands/celebrities} if they are also mentioned in the provided annotation
    \item The sentence \underline{} (e.g., “I should buy this perfume because roses are red and violets are blue”) but it \underline{should not be ungrammatical} (do not write something like “I should hello world because rainbow”)
\end{enumerate}

\noindent
Rule 8 was introduced as there is evidence \cite{clipneurons} that CLIP is sensitive to proper nouns. Therefore, we wanted to avoid our negatives being preferred by the model simply because they contained more detailed information. 

Our annotation interface allowed annotators to flag ads in case of:
\begin{enumerate}[topsep=1pt,itemsep=-3pt]
    \item Presence of inappropriate/offensive/harmful content.
    \item Low readability of the text.
    \item Low image resolution.
    \item Being unable to understand the ad (e.g., because the text was not in English).
    \item Being unable to create a distractor meeting all the requirements.
\end{enumerate}

\section{Dataset Examples}
\label{sec:appendix_examples}
\begin{figure*}
    \centering
    \includegraphics[width=\textwidth]{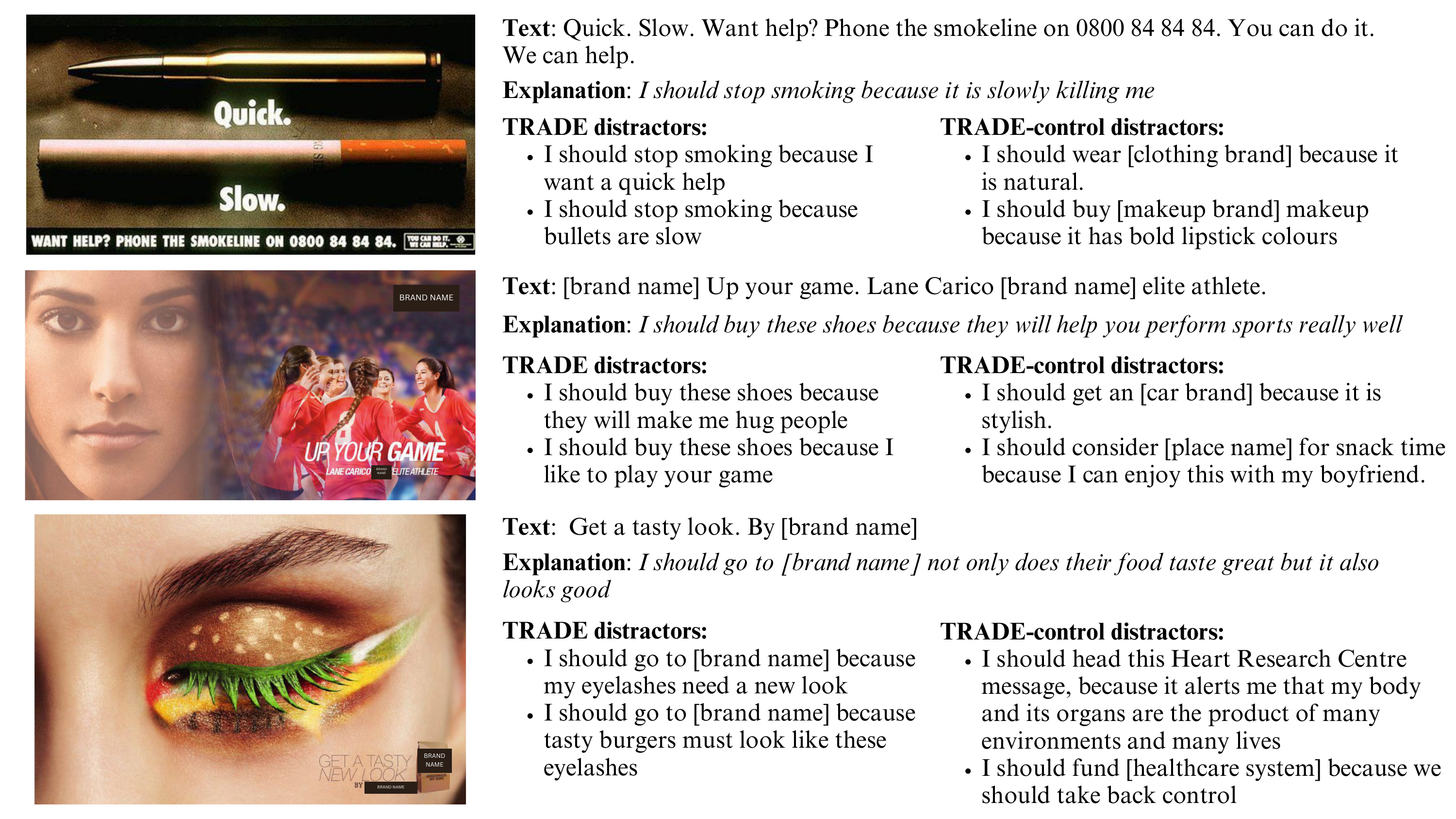}
    \caption{Examples from TRADE and TRADE-control, along with our transcription of the text (just for readability, not part of the dataset). Brands and logos are edited out in the paper examples for presentation purposes but are visible in TRADE.}
    \label{fig:more_examples}
\end{figure*}

Some additional examples of the adversarial explanations we collected are shown in Figure~\ref{fig:more_examples} along with their TRADE-control counterparts.

\section{Human Accuracy on TRADE}
\label{sec:appendix_human_acc}
To quantify the human accuracy on TRADE, we used the crowdsourcing platform Appen to present participants with the ad along with the question ``What should you do according to this ad, and why?'' and 3 options, i.e., a matching explanation and two adversarial grounded negatives. After some unsatisfactory pilot experiments where crowdworkers were not able to pass very simple test questions, we established that the task was not suitable for crowdsourcing. Therefore, we recruited 17 participants who volunteered for the task of judging the 300 samples in TRADE. They were not involved in the creation of the adversarial explanations and were informed that their anonymised data would be included in a study about automatic ad understanding. We ensured all annotators were proficient in English. Each question was answered by 2 different participants. They annotated an average of 35 ads each ($std=14$, $max=50$, $min=10$). The mean accuracy calculated over the 600 collected judgements was 94\%. The cases where both participants selected the target explanation were 270 (90\%). 

\section{Tested Models}
\label{sec:appendix_model_overview}

Here we provide an overview of the models used in our experiments. 

\paragraph{CLIP} \cite{clip} is a contrastive model where image and text are separately encoded by two transformer-based models and then projected to the same vector space. CLIP is trained with a contrastive loss that minimizes the cosine distance between matching pairs of image and text embeddings. We used it in the Hugging Face implementation.\footnote{\url{https://huggingface.co/docs/transformers/model_doc/clip}}

\paragraph{ALIGN} \cite{align} is also a contrastive vision-and-language model trained with the same loss function used for CLIP. It mainly differs from the latter in its encoders (EfficientNet for images and BERT for text) and in that also leverages noisy data during the training process. We used the Hugging Face model implementation.\footnote{\url{https://huggingface.co/docs/transformers/model_doc/align}}

\paragraph{LiT} \cite{lit} is a contrastive model where the image encoder is “locked” (i.e. frozen) during pre-training, whereas the language encoder is initialized with random weights and trained from scratch with a contrastive loss. We used the Vision Transformer implementation\footnote{\url{https://github.com/google-research/vision_transformer}} by Google Research. 

\paragraph{ALBEF} \cite{albef} is a vision-and-language model consisting of two separate transformer-based encoders from image and text and a multimodal encoder. The uni-modal modules are pre-trained contrastively and their outputs are then fused in the multimodal module, which is pre-trained with masked-language-modeling and image-text-matching objectives. We used the LAVIS implementation by Salesforce.\footnote{\url{https://github.com/salesforce/LAVIS}}

\end{document}